\documentclass[letterpaper, 10 pt, conference]{ieeeconf}  % Comment this line out if you need a4paper

\usepackage{graphicx}
\usepackage{algorithm}
\usepackage{algorithmic}
\usepackage{authblk}
\usepackage{cite}

\newlength\myindent
\newcommand\bindent{%
  \begingroup
  \setlength{\itemindent}{\myindent}
  \addtolength{\algorithmicindent}{\myindent}
}
\newcommand\eindent{\endgroup}

\IEEEoverridecommandlockouts                              % This command is only needed if 
\title{\LARGE \bf
Lifelong update of semantic maps in dynamic environments
}

\author{Manjunath Narayana and Andreas Kolling and Lucio Nardelli and Phil Fong% <-this % stops a space
}
\affil{iRobot Corp., Pasadena, USA}

\begin{document}

\maketitle
\thispagestyle{empty}
\pagestyle{empty}

%%%%%%%%%%%%%%%%%%%%%%%%%%%%%%%%%%%%%%%%%%%%%%%%%%%%%%%%%%%%%%%%%%%%%%%%%%%%%%%%
\begin{abstract}
A robot understands its world through the raw information it senses from its
surroundings. This raw information is not suitable as a shared representation
between  the robot and its user.
A semantic map, containing high-level
information that both the robot and user understand, is better suited to be a 
shared representation.
%With repeated sequential runs, as a robot updates its raw map with newly sensed information,
%the semantic map also needs consistent maintenance.
We use the semantic map as the user-facing interface on our
fleet of floor-cleaning robots.
Jitter in the robot's sensed raw map, dynamic objects in the environment, and 
exploration of new space by the robot are common challenges for robots.
Solving these challenges effectively in the context of semantic maps is key to 
enabling semantic maps for lifelong mapping.
First, as a robot senses new changes and alters its raw map in successive runs, 
the semantics must be updated appropriately.
We update the map using a spatial transfer of semantics.
Second, it is important to keep semantics and their relative constraints
consistent even in the presence of dynamic objects.
Inconsistencies are automatically determined and resolved through the introduction of 
a map layer of meta-semantics.
Finally, a discovery phase allows the semantic map to be updated with
new semantics whenever the robot uncovers new information.
Deployed commercially on thousands of floor-cleaning robots in real homes, our user-facing
semantic maps provide a intuitive user experience through a lifelong mapping robot.
%The semantics may include concepts inferred automatically through algorithms or
%annotated through user-input, Real world robots have to account for dynamic environments and noise Semantics annotmaintained by a robot, that include raw information,
%annotations, and constraints between the annotations, need to be 
%maintained consistently over time.
%Noise in the robot's sensing or changes in its dynamic
%environment can alter the raw map, invalidate the annotations, or violate
%the constraints.
%In sequential missions, as the robot updates its raw map with new 
%information from its sensors, a system is required for transferring the
%semantics onto the new raw map to ensure they remain valid.
\end{abstract}

%%%%%%%%%%%%%%%%%%%%%%%%%%%%%%%%%%%%%%%%%%%%%%%%%%%%%%%%%%%%%%%%%%%%%%%%%%%%%%%%
\section{Introduction} 
The raw information map where a robot saves a representation of its sensed information
is essential to the robot's tasks such as planning, navigation, and obstacle avoidance.
However, because it consists of low-level information which is difficult for
humans to comprehend, it is not a suitable representation to use for a shared understanding of 
the world with the user. 
A semantic map on the other hand is a richer representation of the environment
that can be understood by both the robot and the user.
It is better suited to be a common frame of reference and
a user-facing representation.
%The shared reference frame enables the semantic map to be the center piece of 
%an intuitive user interface with the robot.

%For example, in a floor-cleaning robot, once the robot has explored its
%surroundings, it has a reasonable understanding of its environment.
%The raw information, perhaps the map of the occupied pixels where the 
%robot sensed obstacles, may not be easy for a user to understand.
%Processing the raw information and annotating high level structures 
%such as walls and furniture on it is a much better representation 
%to share with the user.
\begin{figure*}[!htb] \centering \includegraphics[width=180mm]{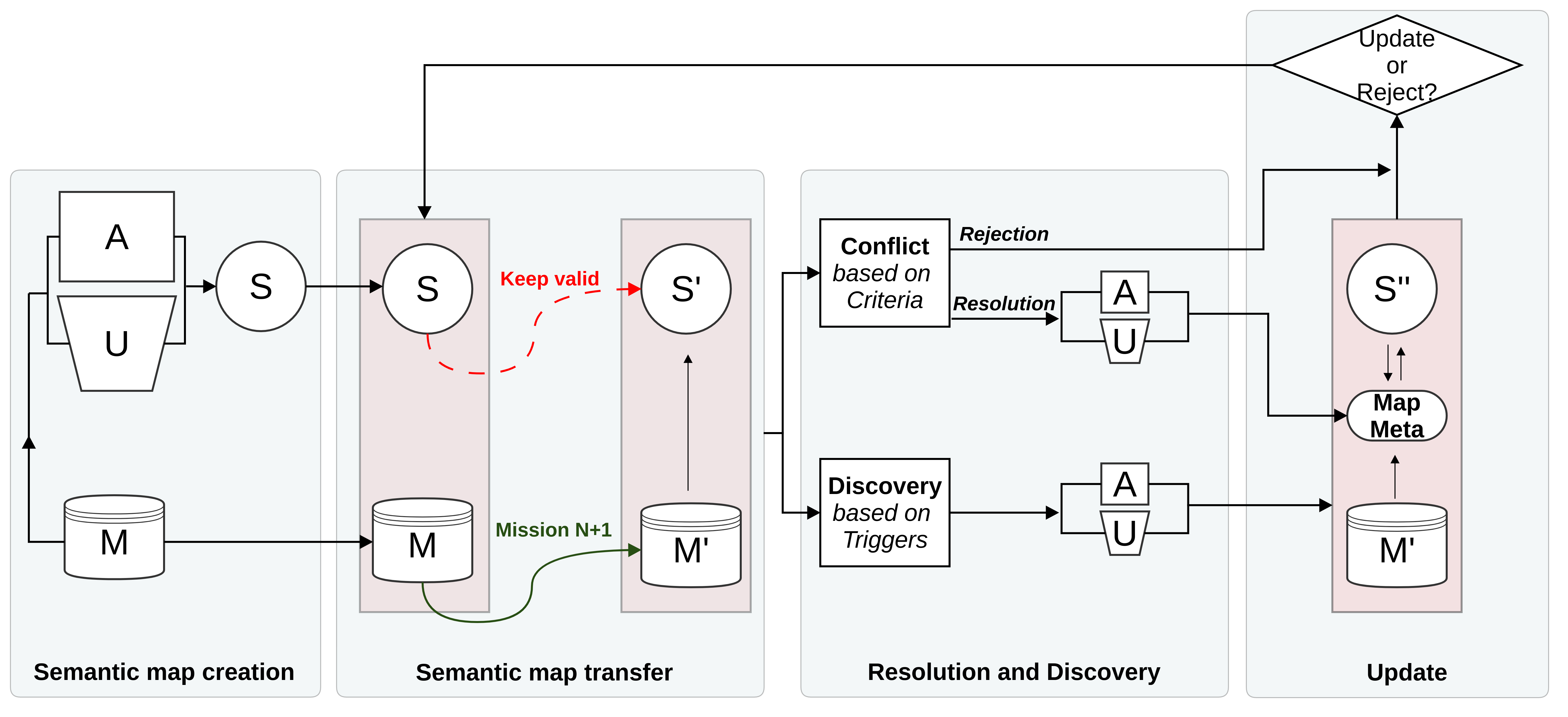}
\caption{\textbf{Semantic map update system.} Semantic map, consisting of semantics S annotated on the raw map M, is created using algorithms A or user input U. Semantic map transfer attempts to transfer all semantics from the previous map N to map N+1 at the end of robot mission N+1 with the requirement that the semantics remain valid. Conflicts in the semantics are resolved and/or new semantics are discovered in the Resolution and Discovery module. Map meta layer is introduced during conflict resolution to keep semantics valid without altering the robot's raw map. Finally the Update module determines whether to update or reject the new information from mission N+1.} \label{transfer_block} \end{figure*}

We employ semantic maps in our fleet of floor-cleaning robots where users
use them extensively to interact with their robots.
Our robots use a visual SLAM (vSLAM \cite{c22_Eade10}) localization system to 
generate and maintain a lifelong map \cite{c24_Banerjee19} of its environment.
Our semantic map which consists of walls, doors, and rooms of the home
is generated through algorithms \cite{c25_Kleiner17} and user annotations.
The semantic map is central to the user's interactions with the robot. 
It enables intuitive experiences such as cleaning a 
specific room through an app or voice command.

Deploying semantic maps on a large fleet of robots exposed
several common challenges in robot mapping that needed to be addressed
in the context of semantic mapping.
Noise in the sensing, dynamic objects in the environment, and
the robot uncovering new space that it had not previously explored are 
three such problems that occur regularly in practical robots.
We present solutions for these specific problems through a framework that is 
general enough to handle other similar problems.  

In real-world situations, under noisy sensing and 
dynamic environments, the low-level information
may be sensed differently between different robot runs, whereas the high-level semantics 
of the environment are not expected to change significantly.
As a user-facing representation, it is desirable that the semantic
map is stable across slight perturbations of the underlying raw map, 
but flexible to account for significant and semantically meaningful
changes in the environment.

The semantics, once established, are the common frame of reference 
for the robot and user.
For instance, the user can intuitively communicate to the robot 
to clean a particular room and the robot would know exactly what to do.
It is essential that once semantics have been exchanged between the user
and the robot, they remain valid for the lifetime of the robot. 
Semantic maps have been discussed extensively in the literature \cite{c1_Kostavelis15, c7_Landsiedel17, c8_Nuchter08, c9_Ruiz17, c13_Galindo05}.
They can result in better understanding of indoor maps \cite{c2_Rusu09, c3_Pangercic12, c4_Zender08} 
and outdoor spaces \cite{c5_Lang14, c6_Wolf08}, and can be useful in tasks such as 
planning \cite{c10_Galindo08}.
While several works address the challenges in semantic maps in a single
robot run, relatively fewer have focused on multiple sequential runs \cite{c11_Mason12, c12_Galindo07},
which is a requisite for lifelong mapping.

When a lifelong mapping robot performs a new run (called \emph{mission}), 
it is normal to update its raw map using 
recent information from its sensors. 
Every raw map update must be accompanied by an update of the semantic map.
The advantage of updating the raw and semantic map every mission is that
the robot maintains the most recent belief of its environment.
For lifelong mapping, localization in dynamic environments \cite{c19_Tipaldi13}, efficient 
management and update of maps \cite{c15_Brki18, c17_Pomerleau14}, and map summarization \cite{c16_Dymczyk16, c18_Muhlfellner16}
have been studied previously.
The challenges in ensuring valid semantics for lifelong robots have not been
not addressed.

Semantic constraints \cite{c14_Limketai05, c13_Galindo05}, particular dynamic ones \cite{c20_Asada89}
are an important aspect of maps.
Semantics are often constrained by how they relate to each other and
to the raw map. An example constraint in our system is that doors must be
attached to a wall.
Unsuccessful update of semantics implies that there are
inconsistencies (or \emph{conflicts}) in the constraints.
For instance, a door which previously separated two rooms may 
be inconsistent by appearing between two different rooms after an unsuccessful update.
When semantics are updated incorrectly, the assumption that the robot and user 
share a common understanding of the environment is violated.
%The consequence of incorrect transfer is that the robot 
%might behave differently from the user's expectations.
%For example, if a room boundary is corrupted by becoming too small, 
%the robot would clean only part of the room which is not 
%the user's intent.
We use algorithms to detect conflicts and resolve them.
For conflict resolution, we propose the use of an additional map layer called map-meta layer
with some additional meta-semantics.
Use of multiple-layered maps is not new in the literature. For instance, Zender et al. \cite{c21_Zender07}
use multiple layers, with 
each layer representing a different level of abstraction.
We use the additional map layers for 
explicitly handling the inevitable inconsistencies over sequential map updates.

If the semantic map update is
not successful despite the meta-layer based resolution, the map update may be discarded
and the map reverted to the valid state from the previous mission.

Apart from maintaining previously established semantics,
when the robot senses significant changes in the environment, 
there is potential for discovering new semantics.
A discovery step enables computation of these semantics and their annotation 
on the map.

Figure \ref{transfer_block} shows a block diagram of our system
that consistently updates the semantic map across multiple missions 
while allowing for the robot to learn and incorporate environment changes 
into the semantic map. 
We motivate and illustrate the semantic map system
through our use-case of floor-cleaning robots.
Our semantics are derived on simple 2D floorplan maps, but we believe the
building blocks and principles described in this paper are useful 
for other complex maps that involve semantics and constraints.
The internal algorithms for the building blocks are 
explained at a high level with less emphasis on the details
because the exact algorithms will vary greatly depending 
on the robot's type and its purpose. 

The paper is organized as follows.
Section \ref{Overview} describes our robot's localization and mapping system,
semantics and constraints, and the various modules of the semantic map update system.
The problem of jitter in robot's sensed map is addressed by our algorithm 
to transfer semantics from mission to mission in Section \ref{Transfer}.
Dynamic objects can lead to different types of semantic conflicts.
Conflicts, their automatic detection, and resolution using meta-semantics 
are explained in Section \ref{Conflicts}.
Section \ref{Discovery} handles discovery of new space and finally Section \ref{Update}
recaps the update step of the semantic map system.
Quantitative results in Section \ref{Results} show
our system's efficacy on hundreds of real home maps.

\section{Semantic map system overview}\label{Overview}
\subsection{SLAM to Semantic map}\label{SemanticMap}
Our robots use a visual SLAM \cite{c22_Eade10} (vSLAM) algorithm that 
tracks visual landmarks and integrates
various sensor modalities to generate a map of the house.
Once generated, the vSLAM map is updated for the lifetime of the robot 
across hundreds of missions using efficient
techniques for managing the complexity while retaining its efficacy \cite{c24_Banerjee19}.

\begin{figure}[!htb] \centering \includegraphics[width=70mm]{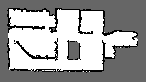}
\caption{Raw occupancy map in robot} \label{occ} \end{figure}

The vSLAM map is crucial for the robot to localize itself accurately 
at different locations and times in the home. Another important map in 
the system is the \emph{Occupancy map} that represents a top-down view of the 
home where white represents \emph{free} areas,
black represents \emph{occupied} areas where walls or objects 
were sensed (called \emph{obstacles}), and grey represents \emph{unexplored} regions.
This map is saved internally through a set of sub-grids that are inter-connected
through pose constraints and are allowed to move across each other\cite{c23_Llofriu17}.
The pose constraints come from the vSLAM algorithm and hence there is a
tight coupling between the occupancy map and the vSLAM localization system.
Figure \ref{occ} shows an example of our two-dimensional occupancy map.

\begin{figure}[!htb] \centering \includegraphics[width=70mm]{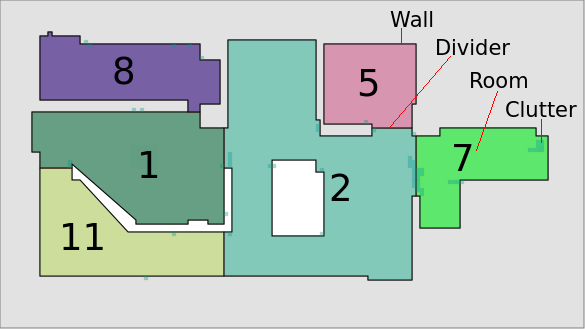}
\caption{Semantic map derived from raw map of Fig \ref{occ}}
\label{map_1} \end{figure}

Because the obstacles are detected by a bump sensor of a limited resolution and 
due to uncertainty in a robot's estimate of its own pose, the occupancy map can be noisy and
difficult for a user to comprehend. Extraction of walls, dynamic 
objects (called \emph{clutter}), rooms,
and dividers from the occupancy map yields a semantic map of 
Figure \ref{map_1}, which is much easier to understand. 
Shown as \textbf{Semantic map creation} module in Figure \ref{transfer_block},
the semantics are both estimated automatically \cite{c25_Kleiner17} and annotated directly
by the user through an app interface.

%These semantics, their meaning, and their relative constraints are shown in 
%in table \ref{Semantics_table}.
%These semantics are merely examples - their meaning and constraints 
%are specific to our robot and its applications. The motivation for our 
%chosen constraints are listed in the last column.
%Other robots may define semantics and constraints specific to their applications.

%The \textbf{Semantic map creation} block in Figure \ref{transfer_block}
%shows that the semantics may be extracted using
%algorithms $\mathbf{A}$ or user input $\mathbf{U}$. 
%The raw map $\mathbf{M}$ and the semantics $\mathbf{S}$ together 
%constitute the semantic map. 

\begin{figure}[!htb] \centering \includegraphics[width=70mm]{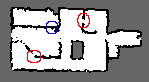}
\caption{New raw map after a subsequent mission. Compared to earlier map in Fig \ref{occ}, previously sensed walls appear disconnected (highlighted in red) and an opening has closed (blue)}
\label{occ_current_marked} \end{figure}

\begin{figure}[!htb] \centering \includegraphics[width=70mm]{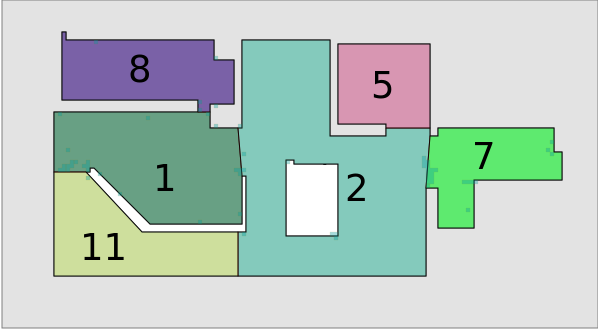}
\caption{Updated semantic map. Semantics from Fig \ref{map_1} successfully transferred to raw map in Fig \ref{occ_current_marked}}
\label{map_1_transfer} \end{figure}

\begin{figure}[!htb] \centering \includegraphics[width=70mm]{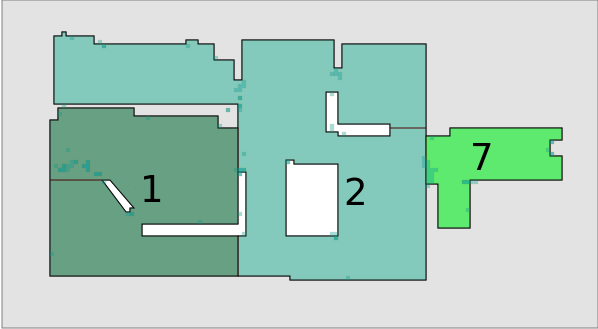}
\caption{Semantic map transfer failure. Failed attempt to transfer semantics from Fig \ref{map_1} to raw map in Fig \ref{occ_current_marked}. The changes highlighted in Fig \ref{occ_current_marked} cause rooms $5$,$11$, and $12$ to not be recovered correctly.}
\label{map_1_transfer_fail} \end{figure}

\subsection{Multiple missions and the Semantic map}\label{MultiMissionSemanticMap}
After the semantic map has been created, 
in subsequent missions, the robot may sense its world differently because of changes in 
the world or uncertainty in the robot's representation of the world.
Further the vSLAM system constantly updates the robot's position 
and improves its belief about its world as it observes visual information. 
Hence even pixels representing static objects may move slightly in the 
occupancy map to reflect this belief.
For instance, Figure \ref{occ_current_marked}
shows the occupancy map of the same space as Figure \ref{occ},
but at the end of a different mission. 
%Note that the occupancy map is incremental and shows both historical
%as well as the most recently sensed obstacle pixels.
The red and blue ovals show the key changes
sensed by the robot.
The red ovals show previously connected obstacles now 
disconnected either due to moved objects or  
sensing errors.
The blue oval shows a previously open path closed.
Here, the robot sensed obstacles where the 
previous doorway was and decided that it can not enter the room.

Although the underlying raw map varies slightly with each mission, 
the user is generally not interested in these perturbations because
semantic concepts tend to be stable in a home.
We would hence like to show the user a stable semantic 
map by transferring the semantics correctly from the previous map
to the new one.

Figure \ref{transfer_block} shows that as a 
robot performs a new mission, it updates its internal map ($\mathbf{M^\prime}$). 
The requirement for the \textbf{Semantic map transfer} block is that the semantics are updated ($\mathbf{S^\prime}$)
while remaining valid. 
Successful transfer of semantics is shown in Figure 
\ref{map_1_transfer}.
All the rooms and dividers are reproduced correctly in the transferred map.
An unsuccessful transfer attempt shown in 
Figure \ref{map_1_transfer_fail} is undesirable because some rooms 
have been lost and others have changed their shape
significantly.

\begin{table*}[h]
\caption{Semantics for indoor robots}
\label{Semantics_table}
\begin{center}
\begin{tabular}{|c|c|c|c|c|}
\hline
\textbf{Semantic} & \textbf{Physical meaning} & \textbf{Constraint} & \textbf{Motivation for constraint}\\
\hline
\hline
Occupancy pixels & Occupied regions in map & Must be sensed by robot & Used as cost map in planning, constraint\\
                                                                 & & & helps use robot's best estimates for occupied pixels\\

\hline
Clutter & Temporary/dynamic objects & Lies on occupancy pixels & Yields accurate visualization on semantic map\\
\hline
Wall & Shape of environment & Within a distance of occupied pixels & Yields accurate representation of rooms\\
\hline
Divider & Boundary between rooms & Ends on wall or another divider & Enclosed rooms are well-defined \\
        &                        & Annotated by user               & User is final authority on segmentation of the space\\
\hline
Room & Spaces in the environment & Defined by wall sections and dividers & Room polygon tightly corresponds to physical space,\\
     &                           &                                       & hence robot can cover space satisfactorily\\
\hline
\end{tabular}
\end{center}
\end{table*}

\subsection{Semantic constraints}\label{Constraints}
The semantic map for our floor-cleaning robot is designed with several constraints.
While some constraints reflect practical limitations of the robot's application, 
others ensure consistency in the semantic map's visualization.
Table \ref{Semantics_table} lists the semantics and their constraints in our semantic map.
To emphasize the significance of constraints, let us consider the constraint enforced
on the \emph{room} semantic. 
A room is an enclosed space in the home, defined by \emph{walls} and \emph{dividers}.
Walls are constrained to lie within a threshold distance of a sensed occupancy pixel 
and dividers must end on a wall or on another divider. This means that the room polygon thus
described tightly fits the physical walls of the home as 
sensed by the robot. 
%We choose this definition of a room because it gives our robots the best chance to clean
%the physical space as intended by the user. 
%Alternative definitions of rooms are possible - for example, a room may 
%be defined to be a simple rectangle at a given location in
%the 2D map. This alternative definition may be suitable for many purposes, but 
%fails to communicate a well-defined region to a floor-cleaning robot.

\subsection{Conflict resolution and discovery}\label{ConflictResolutionDiscovery}
Each semantic map update must be guaranteed to maintain all the defined constraints.
If semantics become invalid because of a constraint violation, 
this is automatically determined as a \textbf{Conflict} 
using various criteria in Figure \ref{transfer_block}. 
Conflict resolution is done either through 
algorithms or requesting user input.

To aid conflict resolution, we add an additional layer 
called \textbf{Map Meta} to the semantic map. The role of the
meta-semantics layer is to add or remove relevant pieces of information 
that make the semantics valid while keeping the raw map unaltered.
When valid resolution is not possible, the map update is rejected and the semantic
map reverts to the earlier state.
When user input is requested for conflict resolution, the user can either 
provide additional information to fix the map or reject the update.

At the end of each mission, the robot also determines if the map has changed 
significantly and if there is a possibility of annotating new semantics.
This \textbf{Discovery} phase in Figure \ref{transfer_block} enables new semantics to be added
through algorithms or user input.

Next, the general solutions for the three common problems of jitter, 
dynamic objects, and new space are described.

\section{Transfer of map semantics}\label{Transfer} 
\begin{figure}[!htb] \centering \includegraphics[width=70mm]{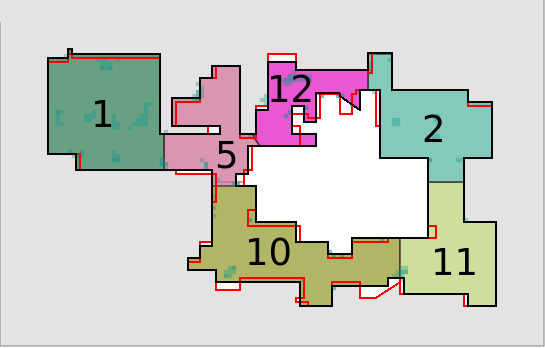}
\caption{Misalignment between tracked room boundaries and new boundary in a map. The new boundary, estimated from the new raw occupancy map is shown in black and is rendered on top of the tracked room boundaries rendered in red.Jitter in the map is evident in the red segments.}
\label{basic_transfer} \end{figure}

The first key problem in dynamic environments is to keep the semantic map
stable even though the underlying raw map may jitter.
The problem can be viewed as one of \emph{transferring} the semantics
from the previous raw map to the current raw map.
Note that our robots already have a very strong vSLAM framework for tracking 
visual landmarks in the home across missions. 
Further, our semantics have a spatial component by virtue of being represented
on a 2D map. 
A natural solution is to leverage the vSLAM framework with modifications
to track the spatial components of the desired semantics.
At the beginning of a robot's mission, 
we integrate the vertices of the rooms and the dividers
into the vSLAM system as explicit spatial points to track.
%(TODO reference).
As the robot moves and updates its location and map,
the locations of the rooms and dividers get tracked by 
the vSLAM system. The room vertices and dividers move
according to the motion estimated by the underlying vSLAM system.
At the end of the mission, vSLAM returns motion estimates for the
rooms and dividers. It is not sufficient to use these estimates as the final 
result since the current shape of the rooms or walls may have changed 
due to removal and adding of clutter objects to the wall, opening and closing
of doors, or moving of large pieces of furniture.
Further, the semantic constraints we defined in table \ref{Semantics_table} for walls
required them to be within a threshold distance of sensed occupied pixels.
Similarly, rooms boundaries were constrained to consist of wall sections and dividers.
In order to guarantee these constraints,
we first estimate the new walls from the occupancy map,
associate the previously tracked divider end-points to the new walls, and then 
reconstruct the room shape from the new walls and dividers.
Figure \ref{basic_transfer} shows the tracked boundaries and dividers visualized
underneath the final boundaries and dividers in an example map.
It shows the inaccuracy in the tracked
room shape and dividers - they do not line up exactly with the walls in the new map. 
Divider end-point association and room reconstruction are hence essential for 
maintaining the specified constraints.

It is also important to keep wall and clutter semantics consistent across maps.
We achieve this by using the tracked room shapes again. 
Obstacles in the new occupancy map that are within a threshold distance 
of the tracked room boundaries are labelled as wall obstacles.
Obstacles that lie in the interior of previously tracked room boundaries are 
labelled as clutter obstacles. 
Wall estimation is done on the new map after this wall and clutter
classification of obstacles.
The result is that walls and clutter are consistent between the previous and 
new maps.

\begin{algorithm}
\caption{Semantic map transfer algorithm}
\label{transfer_alg}
\begin{algorithmic}
\STATE \textbf{1} Track room vertices and divider end-points by integrating them into the SLAM system
\bindent
\STATE (A new occupancy map and tracked room boundaries are the result at the end of the robot's new mission)
\eindent
\STATE \textbf{2} Wall information transfer:
\bindent
\STATE Classify obstacles that are within certain distance 
\STATE to tracked room boundaries as wall obstacles
\eindent
\STATE \textbf{3} Clutter information transfer:
\bindent
\STATE Classify obstacles that are in the interior 
\STATE of tracked rooms as clutter obstacles
\eindent
\STATE \textbf{4} Estimate new walls from the new occupancy map, using the classification from steps 2 and 3
\STATE \textbf{5} Obtain new divider end-points by moving the tracked end-points to the nearest wall
\STATE \textbf{6} Reconstruct new room shapes from the new walls and divider end-points
\end{algorithmic}
\end{algorithm}

\section{Semantic map conflicts}\label{Conflicts}
\begin{figure}[!htb] \centering \includegraphics[width=25mm]{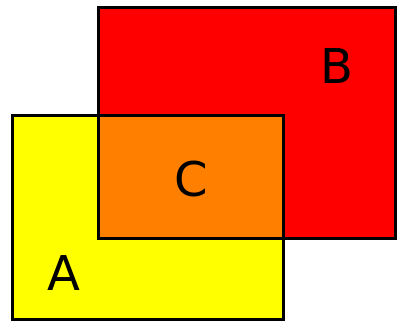}
\caption{Precision-Recall for a single room. Yellow room from previous map is compared
to the red room in the current map. The overlap region is shown in orange.
Recall is $\frac{area(C)}{area(A) + area(C)}$ and 
Precision is $\frac{area(C)}{area(B) + area(C)}$}
\label{PR} \end{figure}

The transfer algorithm \ref{transfer_alg} can result in a semantic map  
with conflicts.
We use precision and recall of rooms to determine when the transfer is successful. 
Figure \ref{PR} illustrates the precision and recall for a single room.
We compute the precision and recall for each individual room in the map. 
Semantics transfer for a given mission is deemed successful, if all the rooms in 
the previous map have been transferred to the new map with precision and recall higher
than a $50\%$ threshold.
Several other criteria may be used, such as failure to transfer all dividers, large unexpected 
change in room shape, and change in connectivity between rooms. 

\begin{figure}[!htb] \centering \includegraphics[width=85mm]{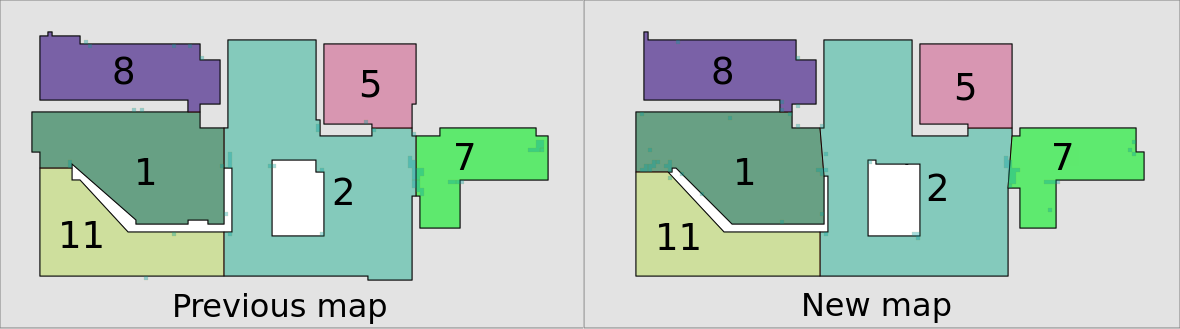}
\caption{Successful transfer with high precision-recall of all rooms}
\label{success} \end{figure}

\begin{figure}[!htb] \centering \includegraphics[width=85mm]{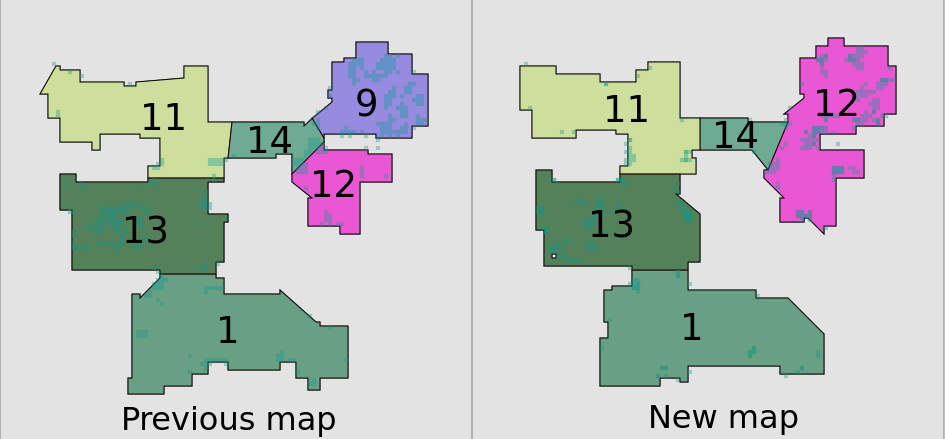}
\caption{Failure case 1 - low recall for room 9 (zero recall), low precision for room 12}
\label{fail1} \end{figure}

\begin{figure}[!htb] \centering \includegraphics[width=85mm]{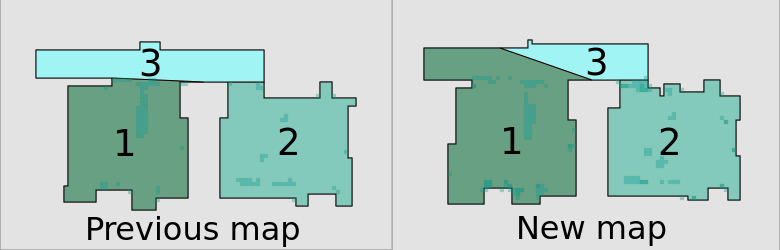}
\caption{Failure case 2 - low recall for room 3}
\label{fail2} \end{figure}

Successful semantics transfer is shown in Figure \ref{success}
where all rooms from the previous map on the left have a high precision 
and recall in the transferred new map.
Figures \ref{fail1} and \ref{fail2} show examples of failures. 
%In Figure \ref{fail1}, the top-right room in the previous map is missing in the new map.
%In Figure \ref{fail2}, even though all the previous map rooms are present,
%the room on top is not recalled sufficiently in the new map.

%TODO - Add PR values for each room in figures

%TODO - Remove if no space in paper

\subsection{Conflict resolution and the need for meta-semantics}\label{Resolution}
The second key problem is of dynamic objects in the home. As objects move around in between 
missions, they can cause significant changes in the occupancy map.
The constraints defined for walls and room boundaries mean that 
when the sensed occupancy pixels differ greatly between missions, it may not be possible
to satisfy all constraints.
One solution is to alter the robot's raw map so that
the semantics may be recovered.
However this is not desirable because the raw map has implications for low-level 
robot behavior and planning.

We propose introducing an additional layer of meta semantics for conflict resolution.
Represented as Map Meta in Figure \ref{transfer_block}, it
enables the recovery of semantics while keeping the raw map unaltered.
Following are some examples of common sources of conflict and recovery.

\subsubsection{Disconnected walls}\label{disconnected_walls}
The previous map's walls may be broken up into disjoint sections in the new map.
When this happens, despite correct placement of the
previous dividers, the new rooms cannot be reconstructed correctly from the 
walls and dividers.
To solve this problem, we take the difference between the tracked room boundaries 
from algorithm \ref{transfer_alg}
and the newly estimated walls. The difference is processed to obtain sections
that were wall in the previous map and are not wall in the new map.
Adding back these connected components to our wall estimation algorithm
makes the wall connections reappear in the new map.
The proposed corrections are biased towards relying on information from the previous
map. 
%This is justified by the fact that the previous map has been 
%validated by the user either explicitly through user annotation or implicitly
%by the user viewing the map.

\begin{figure}[!htb] \centering \includegraphics[width=70mm]{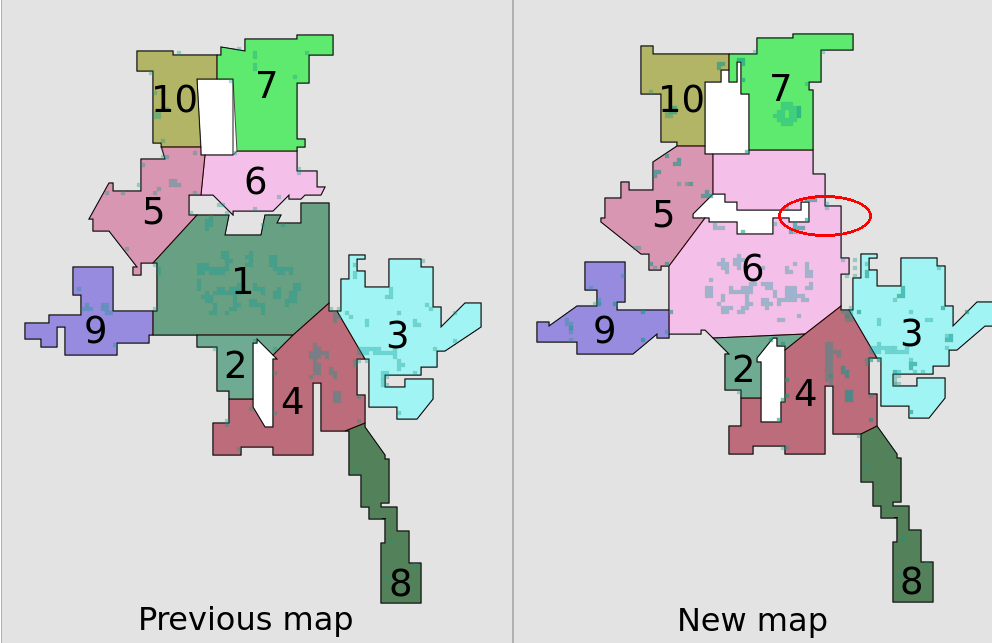}
\caption{Disconnected wall. Room $1$ from previous map cannot be
recovered in new map}
\label{disconnected} \end{figure}

\begin{figure}[!htb] \centering \includegraphics[width=70mm]{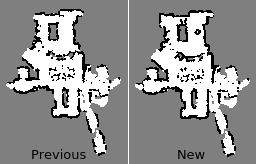}
\caption{Disconnected occupancy map. Underlying cause for disconnected wall in Fig \ref{disconnected}}
\label{disconnected_occ} \end{figure}

Figure \ref{disconnected} shows the previous map and the result
of semantic transfer. The walls that got disconnected (highlighted in red) cause the
room shapes to not be reconstructed correctly.
The underlying cause of the disconnections is apparent in the corresponding raw occupancy 
maps shown in Figure \ref{disconnected_occ}.

\begin{figure}[!htb] \centering \includegraphics[width=85mm]{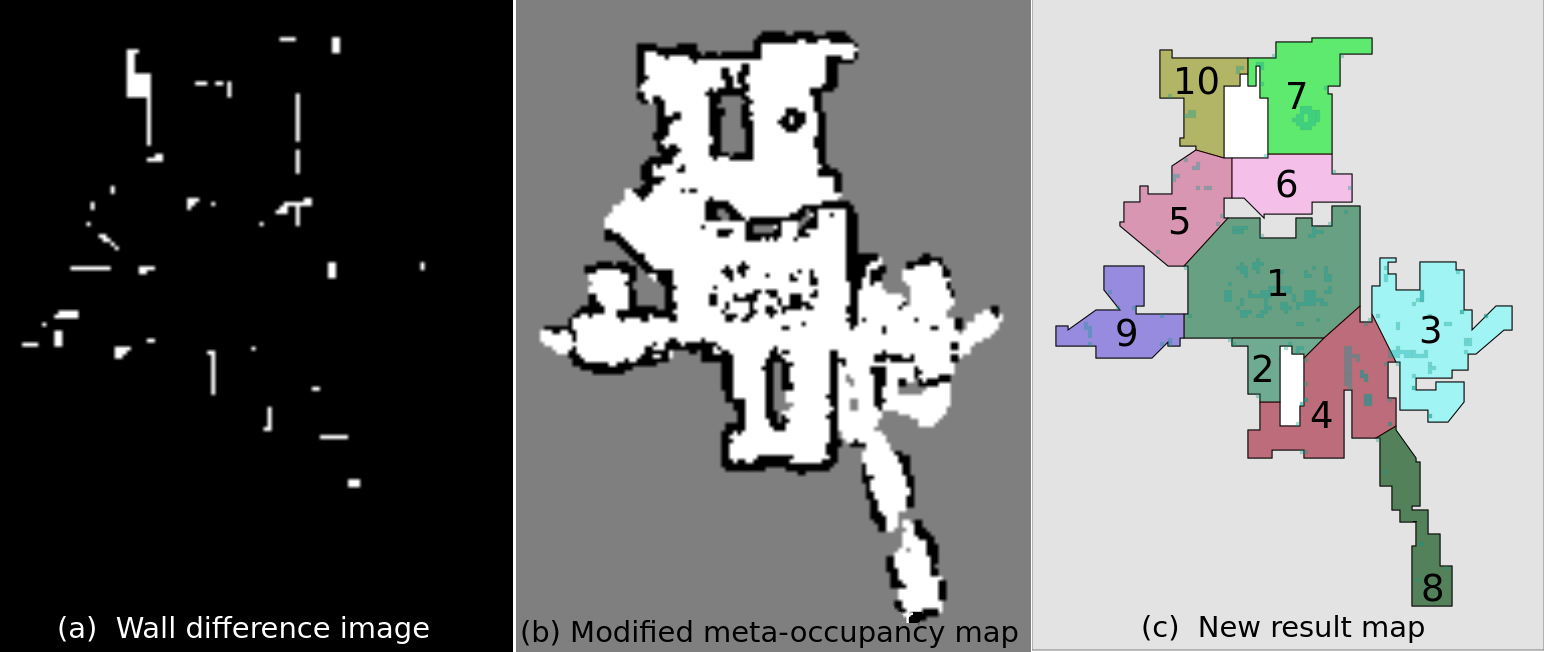}
\caption{Disconnected wall correction. Adding the \emph{wall difference image}
(left) to obtain the modified meta-occupancy map (middle) corrects the disconnected wall in the result semantic map (right)}
\label{wall_difference} \end{figure}

By adding back the wall difference image shown on the left in Figure \ref{wall_difference} at
appropriate sections, we recover a connected occupancy map shown in the center.
Only difference sections that adjoin failed rooms are added back.
Performing wall estimation and semantics transfer on this corrected occupancy map results 
in the correct result on the right. 

\subsubsection{Newly connected walls}\label{new_connected_walls}
This is the converse problem to disconnected walls - two wall sections that were disconnected
in the previous map appear to be connected in the new map. The solution is the converse - 
we take the difference image between the free space within the tracked room boundaries and
the free space within the new walls.
The connected components from this difference image represent sections that are free in
the previous map but wall in the new map.
Adding back these free sections to the occupancy map yields the original openings between walls.

The corrected occupancy map and difference sections are not saved directly in the robot's occupancy pixels,
but as meta-occupancy pixels in the Map Meta layer.
Note that adding back all the wall and free difference sections estimated between the
previous and new map will result in a corrected occupancy map that is identical to the 
previous occupancy map. This would mean that the robot disregards any newly sensed
information from the environment and is not desirable. 
We avoid this by only adding difference sections 
that adjoin rooms that failed to transfer and only if the difference sections are smaller 
than a pre-determined threshold.
Thus, relatively small corrections are made to the map. Large changes sensed by the robot are 
always kept as is.

\subsubsection{Additional dividers}\label{separators}
\begin{figure}[!htb] \centering \includegraphics[width=70mm]{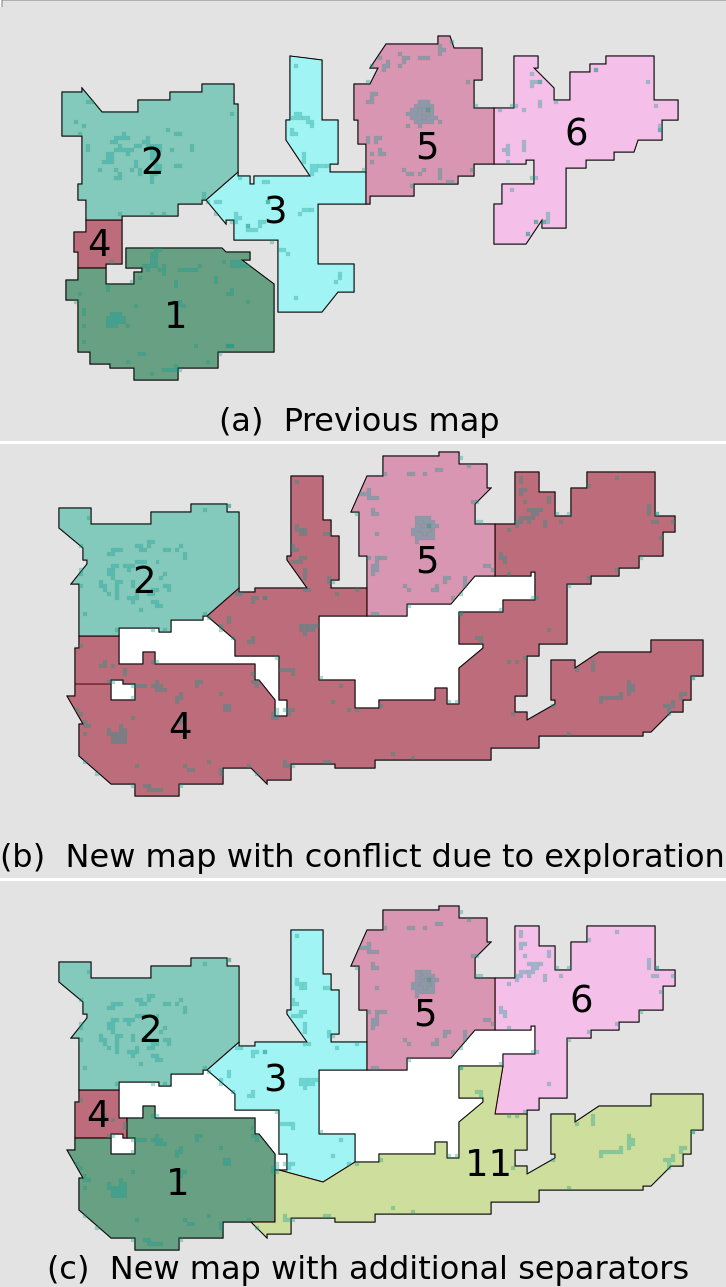}
\caption{Adding additional separators. New space is explored by the robot in (b) that 
connects rooms $1$, $3$, and $6$, resulting 
in their not being recovered by the semantic map transfer algorithm. Adding additional
separators that are not present in (a) allows for recovery of all the original rooms and
creation of new room $11$ in (c)}
\label{map_grow_conflict} \end{figure}

The third key problem mentioned earlier was that of new space being explored 
by the robot.
Sometimes it becomes necessary to add additional dividers to keep the previous and 
new map's room boundaries consistent.
For example, imagine that a new passage connects 
previously disconnected rooms in a home.
An example of this is shown in Figure \ref{map_grow_conflict}. 
In a new mission, the robot explored 
and found a connecting space between previous rooms.
Since this region was unexplored in the previous map, the difference information 
of Sections \ref{disconnected_walls} and \ref{new_connected_walls} does not help. 
The new connection between the previous rooms causes ambiguity for the robot.
This conflict can be detected automatically and can be solved either
through user input or by automatically placing additional meta-dividers.
Meta-dividers are placed by determining which rooms were lost and adding back all 
the tracked boundary segments of the lost room.
The meta-divider based room recovery shown in the bottom of Figure \ref{map_grow_conflict}
shows successful recovery of all the previous room shapes
and inference of a new room.

\begin{table*}[h]
\caption{Meta semantics with relaxed constraints}
\label{meta_semantics_table}
\begin{center}
\begin{tabular}{|c|c|c|}
\hline
\textbf{Meta-semantic} & \textbf{Relaxed constraint} & \textbf{Corresponding semantic (and constraint)}\\
\hline
\hline
Meta-occupancy pixels & Need not be sensed by robot & Occ. pixels ( Must be sensed by robot)\\
\hline
Meta-divider & Ends on wall or meta-divider & Divider (Ends on wall or divider, \\
             & Not annotated by user               &   must be annotated by user) \\
\hline
\end{tabular}
\end{center}
\end{table*}

\subsection{Map Meta layer}\label{MapMeta}
Meta-semantics added to the map meta layer help re-establish the original
set of semantics.
They hence have a form that is similar to the original semantics. 
However, since they need to explain an underlying conflict, 
the constraints on them have to be relaxed.

For example, we added meta-occupancy and meta-divider semantics 
to the Map Meta layer.
The meta-occupancy pixels are a version of the occupancy pixels, but the requirement
that the occupied regions represent real obstacles is not met.
A meta-divider is like a divider. But unlike a 
divider which has to have been annotated beforehand at specified regions of the map, it can be drawn 
anywhere as required.
%The actual representation of the Map Meta layer can be varied as long as the 
%underlying raw map remains unaltered. In our example, we store the
%meta-occupancy pixel regions as a list of polygons that we add to a copy of the raw
%occupancy map. The meta-dividers are saved as a list, separate from the list 
%of original dividers.
The relaxation in the constraints of the Map Meta layer semantics
allows flexibility in the system to resolve conflicts and guarantee
the constraints of the original semantics. 
Table \ref{meta_semantics_table} summarizes our meta-semantics and their relaxed constraints.

\section{Discovery}\label{Discovery}
The third key problem in life-long mapping is exploration for previously unseen space
by the robot. We have seen an example of this problem when it yields conflicts
in Section \ref{separators}.
In general, for a robot that constantly updates its map across missions, 
a mechanism to discover new semantics is essential.
The robot must have the ability to process the change in its map and determine
if there are potentially new semantics to be discovered.
When pre-determined conditions are triggered, our system launches a discovery
phase to gather new semantics from the raw map.
%Semantics that are algorithmically estimated from the raw map are 
%discovered by executing the algorithms.
Discovery of new rooms in the home is one such example.

\begin{figure}[!htb] \centering \includegraphics[width=70mm]{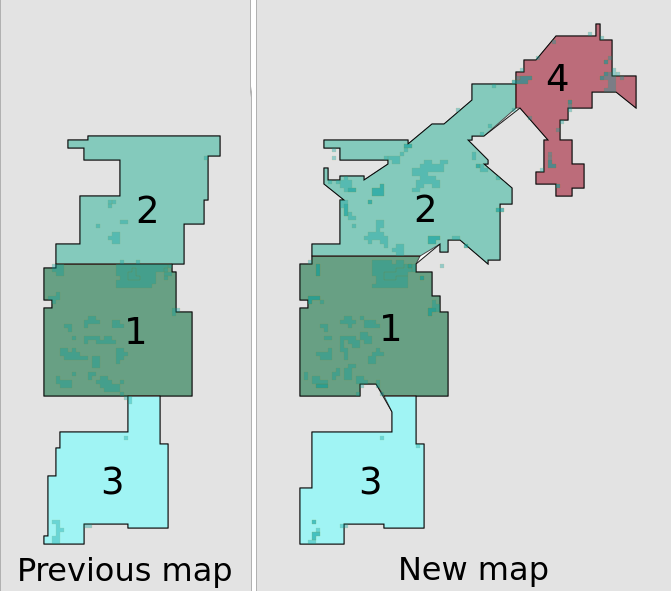}
\caption{New space exploration example. Semantics discovery module detects newly explored space
and infers an additional room $4$ in the new map}
\label{new_space} \end{figure}

When any previous room appears to grow larger than a given threshold in the new map, 
we trigger discovery of new rooms in the map by running an automatic divider estimation 
algorithm in the grown region.
Figure \ref{new_space} shows an example. 
The map at the end of a new mission shows the previous rooms correctly
transferred and a new room added in the newly explored section of the home.
Once the new semantics are discovered, they become available for the robot's use 
and are subsequently maintained by the semantic map system.

\section{Update}\label{Update}
Updating the semantic map with new information allows the robot
to use the information for future missions.
The consequences of updating the robot's semantic map with an invalid one are unexpected
robot behavior and poor user experience. 
To avoid these, the validity of the semantics and their mutual constraints are verified
automatically at the end of the semantics transfer, conflict resolution, and discovery phases.
Only when deemed valid, the new semantic map is set as the robot's map for 
future missions. If found invalid, the robot reverts to the previous map at the
cost of losing information from the new mission.

\section{Results}\label{Results}
The successful use of our semantic maps by thousands of real users
is the most significant result of this work.
We also benchmark our system on a large internal data set from real homes with
cleaning missions performed under no explicit user instructions. 
The data set hence represents realistic usage of a consumer robot.
For quantitative evaluation in this paper, we present results on a small randomly chosen set
of $25$ robots and $425$ missions from our internal data set. 
%, with a distribution of number of 
%sequential missions shown in figure X.
%TODO Add figure describing mission statistics

We use the precision-recall based criterion 
described earlier to determine when a semantic map is 
successfully updated. Table \ref{results_table} shows the error rate - the fraction 
of all missions where the semantic map update failed.
Running the semantic map transfer 
algorithm \ref{transfer_alg} as a baseline results in an error rate of 
$20.92\%$. Including the meta-occupancy semantic and re-estimating walls and clutter
based on the meta occupancy grid reduces the error to $11.06\%$. 
Including the meta-divider semantic along with the meta-occupancy results in
an error rate of $1.41\%$. 
We thus have a highly accurate system for consistent update of semantics across
the multiple missions.  

\begin{table}[h]
\caption{Results}
\label{results_table}
\begin{center}
\begin{tabular}{|c|c|}
\hline
\textbf{Condition} & \textbf{Error rate}\\
\hline
\hline
Semantic map transfer (baseline) & $20.92\%$\\
\hline
Transfer + Meta-occupancy                  & $11.06\%$ \\
\hline
Transfer + Meta-occ + Meta-dividers        & $1.41\%$ \\
\hline
\end{tabular}
\end{center}
\end{table}

\section{Conclusions}\label{Conclusions}

We have described principles for enabling a lifelong mapping robot that 
learns and maintains a semantic map of its world.
Through deployment on a fleet of real robots, 
we uncovered key challenges in semantic maps and their maintenance 
over time in dynamic environments.
Spatial transfer of semantics, automatic conflict detection,
use of meta-semantics for conflict resolution, and
discovery of new semantics form the building blocks of an
intuitive user-facing semantic map system 
that remains stable despite the perturbations in the raw map.
Our semantic map update system addresses the specific key challenges
and serves as a framework for addressing future ones.
%The system components are:
%\begin{itemize}
%\item{Transfer of semantics through spatial tracking}
%\item{Automatic conflict detection}
%\item{Automatic and user-input based conflict resolution}
%\item{Map Meta layer with additional semantics for error recovery while guaranteeing consistent semantics and robot behavior}
%\item{Triggers for detecting the potential or need for new semantic annotation}
%\item{Discovery of new semantics}
%\item{Decision system to update or reject changes to the semantic map}
%\end{itemize}
%We explicitly handle semantics update and conflict resolution through
%use of a meta-semantic map layer. This has the nice property of separating
%the corrections applied from the underlying raw map.
%We have thus described principles for enabling a continuously updating robot that 
%learns and maintains a semantic map of its world.
%Tested on a real-world data set across multiple homes, our system 
%enables an intuitive user experience where the robot adapts its map 
%and semantics to changes in the environment.

\addtolength{\textheight}{-12cm}   % This command serves to balance the column lengths

%%%%%%%%%%%%%%%%%%%%%%%%%%%%%%%%%%%%%%%%%%%%%%%%%%%%%%%%%%%%%%%%%%%%%%%%%%%%%%%%

%%%%%%%%%%%%%%%%%%%%%%%%%%%%%%%%%%%%%%%%%%%%%%%%%%%%%%%%%%%%%%%%%%%%%%%%%%%%%%%%

%%%%%%%%%%%%%%%%%%%%%%%%%%%%%%%%%%%%%%%%%%%%%%%%%%%%%%%%%%%%%%%%%%%%%%%%%%%%%%%%

\bibliography{lifelong_semantic_maps}{}
\bibliographystyle{plain}

\end{document}